\documentclass[sigconf]{acmart}

\usepackage{todonotes}
\usepackage{xcolor}

\AtBeginDocument{%
  \providecommand\BibTeX{{%
    \normalfont B\kern-0.5em{\scshape i\kern-0.25em b}\kern-0.8em\TeX}}}

\setcopyright{acmcopyright}
\copyrightyear{2019}
\acmYear{2019}
\acmConference{DMAIC'19}{August 5}{Anchorage, AL}
\acmBooktitle{DMAIC'19: KDD19 Workshop on Data Mining and AI for Conservation,
  August 5, 2019, Anchorage, AL}
\begin{document}

\title{A Framework for Identifying Group Behavior of Wild Animals}

\author{Guido Muscioni}
\email{guido.muscioni@mail.polimi.it}
\affiliation{%
  \institution{Politecnico di Milano}
  \streetaddress{Milan, Italy 20133}
}

% Change me, riccardo and matteo

\author{Riccardo Pressiani}
\email{riccardo.pressiani@mail.polimi.it}
\affiliation{%
	\institution{Politecnico di Milano}
	\streetaddress{Milan, Italy 20133}
}

\author{Matteo Foglio}
\email{matteo.foglio@mail.polimi.it}
\affiliation{%
	\institution{Politecnico di Milano}
	\streetaddress{Milan, Italy 20133}
}

\author{Margaret C. Crofoot}
\email{mccrofoot@ucdavis.edu}
\affiliation{%
	\institution{University of California, Davis}
	\streetaddress{Davis, California 95616}
}

\author{Marco D. Santambrogio}
\email{marco.santambrogio@polimi.it}
\affiliation{%
	\institution{Politecnico di Milano}
	\streetaddress{Milan, Italy 20133}
}

\author{Tanya Berger-Wolf}
\email{tanyabw@uic.edu}
\affiliation{%
	\institution{University of Illinois at Chicago}
	\streetaddress{Chicago, Illinois 60607}
}

\renewcommand{\shortauthors}{Muscioni, et al.}

\begin{abstract}
  Activity recognition and, more generally, behavior inference tasks are gaining a lot of interest. Much of it is work in the context of human behavior. New available tracking technologies for wild animals are generating datasets that indirectly may provide information about animal behavior. In this work, we propose a method for classifying these data into behavioral annotation, particularly collective behavior of a social group. Our method is based on sequence analysis with a direct encoding of the interactions of a group of wild animals. We evaluate our approach on a real world dataset, showing significant accuracy improvements over baseline methods. 
\end{abstract}

%\begin{CCSXM}
%\end{CCSXML}
%\ccsdesc[500]{TODO ~ Do I need them?}


\maketitle

\section{Introduction}
%A complete observation is critical to understand the complex evolutionary and population processes of wild animals

Understanding animal behavior is central to answering the fundamental question of why do animals (including humans) do what they do. Recently, biologist started to use wearable technologies, such as GPS, accelerometers, and radio sensors, to track animals and their activities. %Motion sensors, such as accelerometers and gyroscopes, are typically used to collect high resolution data about the fine movement of these individuals. 
However, the collected raw data are not human-interpretable and needs to be processed to extract behavioral patterns. The raw data are usually represented as time series that contain timestamped observations of the sensor readings. Biologists also collect behavioral annotations that describe the behavior of an individual (walking, grooming) or a group of individuals (grazing, coordinated movement) during a predefined temporal interval, as well as the context of that behavior (such as the habitat, weather, etc.). In the wild, where the environment cannot be well instrumented, biologists are unable to continuously observe behaviors of wild animals but collect observations over small intervals that are typically insufficient to describe the complex behavioral dynamics of wild animals. Machine learning can help biologists in the process of inferring these behaviors from the sensor data. 

The rise of human wearable technology has lead to the development of new solution to the problem of behavior inference from various sensors. Activity recognition models can be used to learn the relations between the raw time series and the behavioral annotations collected through observations or other modalities. Then, the obtained models can automatically classify the intervals of the collected data for which behaviors were not observed. 

In this work we propose a new framework for inferring group behavior of wild animals from sensor time series. 
%We first outline the works that relate to this paper in section \ref{sec:related}. We provide a formal definition for the problem of behavior inference in section \ref{sec:problem}. We describe the structure of the proposed framework in section \ref{sec:approach}. We then evaluate our work with a real world dataset in section \ref{sec:experiments}. Finally, we discuss the results and provide possible future research directions in section \ref{sec:conclusion}.
\section{Related Work}
\label{sec:related}

The field of activity recognition is directly related to the field of time series classification. There are two main directions to solve this classification problem: based on temporal sequence analysis and using deep learning. 

Temporal analysis methods are based on an explicit description of the raw signals~\cite{banos2014window,lara2012survey}. The temporal stream can be represented as segments~\cite{changpinyo2018multi} or as a whole~\cite{xing2010brief}. The former requires to define the length of the segments while the latter automatically handles the interdependencies between each subsequent observation. In the sequence analysis methods, Conditional Random Fields (CRF) are considered the gold standard~\cite{lafferty2001conditional}, including the specific case of wild animals activity recognition~\cite{li2016adversarial}.

Deep learning methods do not require the explicit description of the raw signals. These methods automatically infer the feature set using the hidden layers. For time series classification, deep learning methods that exploit the power of Long Short-Term Memory (LSTM) components are considered the state-of-the-art~\cite{ordonez2016deep}.
\section{Problem Definition}
\label{sec:problem}
Time series classification problems are characterized by two main components: the selection of the temporal resolution and the selection of the classification model. 

To define these two problems, we first present the basic notation for time series classification in the case of multiple entities.

Let
%\begin{equation}
$\mathcal{U} = <U_1, U_2, ..., U_n>$
%\end{equation}
denote the set of time series of $n$ entities.
Let $U_i$ be a time series of $t_i$ timestamps:
%\begin{equation}
	$U_i = \{{u_i}^1, {u_i}^2, ..., {u_i}^{t_i}\}$.
%\end{equation}
%
Let $\mathcal{L}$ denote the set of labels for each entity:
%\begin{equation}
	$\mathcal{L} =<L_1, L_2, ..., L_{n}>$, 
%\end{equation}
with $L_i$ denoting the set of labels $\{l^j_i\}$, where $i$ is an entity and $j$ is the timestamps for which a label is provided. % that may not be complete

\begin{definition}[Temporal segmentation function]
	A \textit{temporal segmentation function} $\mathcal{F}$ is a mapping between each time series $U_i$ to a set of time windows $\mathcal{W} = \{w_1,w_2,...,w_T\}$,
	where each $w_i$ contains one or more timestamps of $U_i$.
\end{definition}

Given $\mathcal{F}$ and a set of time series $\mathcal{U}$, the classification task is the process to find a model $\mathcal{M}$ such that
%\begin{equation}
	$\mathcal{M}(\mathcal{F}(U_i)) = \{l^1_i,l^2_i,...,l^T_i\}$.
%\end{equation}
The result is a complete label set for each $U_i \in \mathcal{U}$.

\section{The Framework}
\label{sec:approach}
We propose a framework based on sequence analysis, which is composed of two main steps. First we select the best global temporal resolution value. Then, we propose a new way to encode the social relations among a group of entities, which in our case are wild animals.

%\subsection{Temporal resolution selection}
Finding the right temporal resolution is critical for time series classification. We base our approach on an optimization task over each time series in $\mathcal{U}$.
Given a set of possible global temporal lengths, we infer, for each value, a set of consecutive time windows. We validate each inferred set based on one or more metric, which can be interdependently defined for each different classification task.
Given the combined scores of each metric, we select the temporal resolution value that maximizes the total score, thus optimally segmenting the time series.

%\subsection{Social relations encoding}

We directly encode the social intrinsic features of group behavior representing it as social graphs (network). Given the optimally segmented time series, we infer a network over the set of $n$ entities for each segment. %~\cite{ScientificReportsFarine}. 
Then, we extract relevant topological and relational features for a general purpose classification model. 
The network definition is domain-dependent and is assumed to be given (by domain experts) for each  dataset~\cite{,DBLP:journals/corr/BrugereKB17}.
\section{Experiments}
\label{sec:experiments}
%\subsection{Dataset}
{\bf Dataset.} We use a publicly available dataset of group activities of baboons~\cite{Crofoot-1358}, which has been previously used for the activity recognition task~\cite{li2016adversarial}. The dataset contains 26 individuals tracked for 35 days. The labels set contains 8 activities and are available for 2 days only, at a 1 minute (60 sec) resolution. The temporal resolution selection is bounded from below by the label resolution.
For this dataset, the social network was defined based on pairwise proximity. An edge between two baboons exists if they are within 2 meters to each others. This network definition has been proven to be meaningful in the biological research~\cite{strandburg2015shared}. 
For each inferred network we extract topological and relational features, such as the degree of a node, the page rank score and the average of the features of the neighbors.

%\subsection{Experimental setup}
\noindent{\bf Experimental Setup.} We validate our approach using 10-fold cross validation and we report results for accuracy and the weighted F1 score~\cite{sokolova2006beyond}, ensuring same size training and testing instances. As a simple baseline we consider a majority classifier. For state-of-the-art baselines we consider a CRF, the \textit{Adversarial Sequence Tagging} (AST), and a deep learning approach \textit{DeepConvLSTM}.
For our implementation we use XGBoost~\cite{chen2016xgboost} as the classifier.

\section{Results}

\begin{figure}[t]
	\centering
	\includegraphics[width=0.8\linewidth]{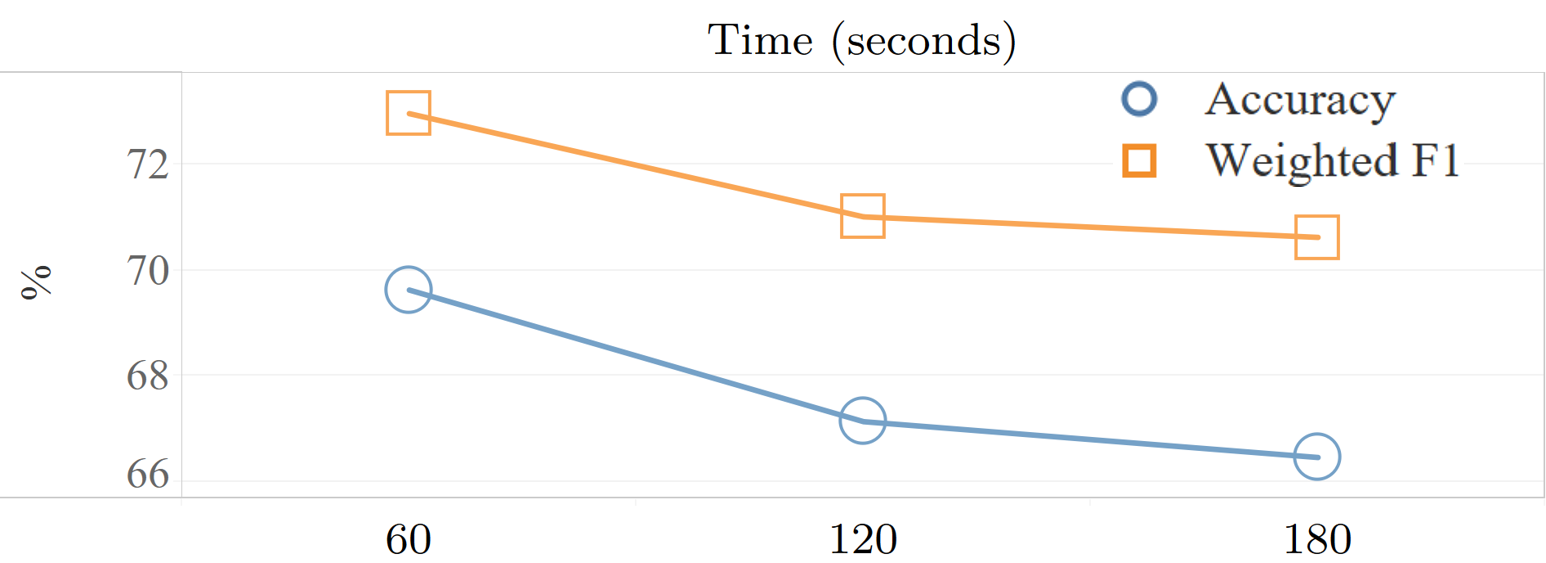}
	\caption{Performance as a function of temporal resolution.}
	\label{fig:grouptrslegend}
	\vspace{-3mm}
\end{figure}

%\noindent{\bf Results.} 
Figure \ref{fig:grouptrslegend} shows the results of the temporal resolution step. The best selected value is 60 seconds (minimum possible) over an interval starting from 60 to 180 seconds. The results shown do not use network information. 
Table \ref{tab:freq} shows the results of comparing our complete approach, including the encoding of the social relations as network information,  with the selected baselines. The metrics show that our framework achieves better performance than the baselines. The lift for the accuracy  is about $10\%$ and  for the F1 score is $6\%$. Adding social information provides a lift of $7\%$  for accuracy and $4\%$  for the F1 over the initial results shown in Figure \ref{fig:grouptrslegend}. %which do not consider them.

\begin{table}[t]
	\caption{Results and comparison with the  baselines.}
	\label{tab:freq}
	\begin{tabular}{lrrrr}
		\toprule
		& Majority & AST & DeepCoLSTM & Our \\
		\midrule
		Acc    & $36.5 \pm0$& $66.05 \pm4.3$    &     $67.50 \pm1.4$         &       $77.88 \pm2.3$          \\
		W-F1 & $17.8 \pm0$ &  $70.05 \pm3.1$   &     $69.88 \pm1.1$         &       $76.54 \pm1.7$          \\
		\bottomrule
	\end{tabular}
	\vspace{-4mm}
\end{table}

\section{Conclusions}
\label{sec:conclusion}
We presented preliminary results for a new framework for inferring group behavior of wild animals. Our approach is based on finding the best possible temporal resolution for classifying the behaviors and an explicit encoding of the social structure of a group of individuals as network information. Our evaluation on a real world dataset shows that the proposed framework better identifies the complex behavioral dynamics of groups of wild animals. 

We are currently working on extending the temporal resolution step to a more dynamic approach allowing varying temporal steps, which will allow to better identify the critical components of each different behaviors. We are also planning to include other datasets.

%\bibliographystyle{ACM-Reference-Format}
\bibliographystyle{abbrv}
\bibliography{ref}

\end{document}